\documentclass{bmvc2k}

\title{Program Generation from Diverse Video Demonstrations}

\addauthor{Anthony Manchin}{anthony.manchin@adelaide.edu.au}{1 2 3}
\addauthor{Jamie Sherrah}{jamie.sherrah@adelaide.edu.au}{1 3}
\addauthor{Qi Wu}{qi.wu01@adelaide.edu.au}{1 3}
\addauthor{Anton van den Hengel}{anton.vandenhengel@adelaide.edu.au}{1 3}

\addinstitution{
The University of Adelaide, Australia
}
\addinstitution{
Apollo Labs, Adelaide, Australia
}

\addinstitution{
The Australian Institute for Machine Learning, Adelaide
}

\runninghead{Manchin et al.}{Program Generation from Diverse Video Demonstrations}

\begin{document}

\maketitle

\begin{abstract}
The ability to use inductive reasoning to extract general rules from multiple observations is a vital indicator of intelligence. As humans, we use this ability to not only interpret the world around us, but also to predict the outcomes of the various interactions we experience. Generalising over multiple observations is a task that has historically presented difficulties for machines to grasp, especially when requiring computer vision. In this paper, we propose a model that can extract general rules from video demonstrations by simultaneously performing summarisation and translation. Our approach differs from prior works by framing the problem as a multi-sequence-to-sequence task, wherein summarisation is learnt by the model. This allows our model to utilise edge cases that would otherwise be suppressed or discarded by traditional summarisation techniques. Additionally, we show that our approach can handle noisy specifications without the need for additional filtering methods. We evaluate our model by synthesising programs from video demonstrations in the Vizdoom environment achieving state-of-the-art results with a relative increase of 11.75\% program accuracy on prior works. 
\end{abstract}

%-------------------------------------------------------------------------
\section{Introduction}
\label{sec:intro}

Inductive reasoning involves using logic to extract general rules from multiple observations and is a skill that is widely viewed as an indicator of intelligence. Humans (and many species of animals) are known to possess the ability to observe demonstrations and subsequently use inductive reasoning to acquire new knowledge and skills without requiring explicit instruction. An example of this behaviour would be children learning to play video games, wherein one is able to observe another player, and learn the rules of the game without having to play the game themselves.

Additionally, children are also able to abstract and generalise information they have induced from one video game and apply that same logical rule set to a completely different virtual domain. An example of this may be that someone observes that a red cross symbol indicates a health related bonus in one games, and subsequently uses that information to infer that similar symbols in a different game are also related to player health. Implying general rules from complex observations however has proven to be a difficult task for machines. In-fact until recently, very little work had been published on extracting general rules from a diverse range of visual observations. Advances in both computer vision and machine learning techniques however are changing this. 
\begin{figure}[t]
    \begin{center}
    \includegraphics[width=\linewidth]{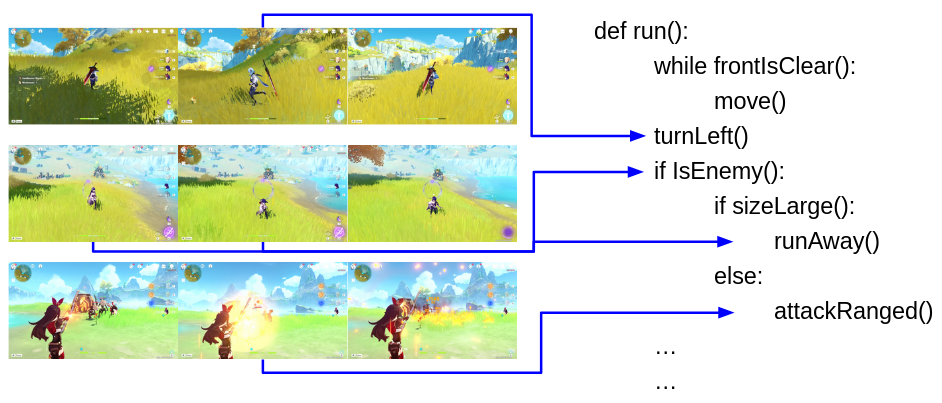}
    \end{center}
    \caption[An illustration of the task of visual program synthesis on the game
’Genshin Impact’]{
    An illustration of the task of visual program synthesis on the game 'Genshin Impact'. Humans can infer a general rule set from simply observing examples of game play. The task of visual program synthesis involves training a machine that can synthesis a program that correctly captures a rule set simply from watching visual demonstrations of agent.
    }
\label{fig:teaser}
\end{figure}

We seek to create a model that can generate executable code by inferring the specifications from multiple visual demonstrations. The goal of creating a model capable of generating executable code has long been a dream of artificial intelligence researchers \cite{ai_dream1,ai_dream2}. However, until recently research has primarily focused on the generation of code, \textit{given} the desired specifications \cite{oracle_specs,syntax_guided}. However, the problem becomes much more difficult when the model is also required to infer the specifications for itself. This task, originally proposed by \cite{demo2program}, presents a unique challenge as it requires a model to accurately detect the relevant semantics of a demonstration, understand the relationships between demonstrations, define a set of specifications that captures this information, and finally generate a program that satisfies these specifications. 

Previous approaches \cite{demo2program,wrc} have considered framing this a sequence-to-sequence task and have utilised the once popular recurrent neural network architecture of long short term memory units as the backbone of their models. However, these approaches were restricted by the limitations of these recurrent based models to handle long sequences. This limiting factor resulted in both \cite{demo2program,wrc} using summarised latent space representations, which undoubtedly limited their models ability to completely capture the entire specification constraints. Meanwhile \cite{PLANS} sought to use a rule-based solver to generate the desired code, which receives its specifications from a neural model. As rule-based solvers are very susceptible to noise, \cite{PLANS} proposed using a dynamic filtering method to de-noise specifications. 

To address these limitations, we propose 'video-to-text transfer transformer' (VT4). We leverage the ability of attention based transformer networks to handle long and disjointed sequences, removing the need to summarise features. Specifically, we formulate a 'visual language' which encapsulates the relevant semantic information from each demonstration. We then simultaneously feed all the demonstrations (represented in a visual language) into our encoder-decoder transformer network which learns to summarise and translate the demonstrations into an executable program. Additionally we show that our approach is able to handle noisy specifications without the need for additional filtering methods. We evaluate our model in a partially observable virtual environment (Vizdoom) \cite{vizdoom} and demonstrate that our approach is able achieve state-of-the-art results by out-performing previous summarisation and rule-based approaches. We observe a relative increase of 15.45\% and 11.75\% over summarisation and rule-based approaches respectively. 

\paragraph{Contributions}
\begin{itemize}
    \item We present video-to-text transfer transformer for the task of executable program generation from video demonstrations. Addressing the limitations of previous summarisation and rule-based approaches, VT4 can generate programs from a long, disjointed set of demonstrations without the need for a summarised representation of the average demonstration. 
    
    \item We evaluate the effects of noisy specifications on program accuracy and show that our model is robust to significant levels of erroneous detections. State-of-the-art results were achieved with a 10\% error rate for perception primitives. Significantly out-performing previous rule-based models which strongly rely on dynamic filter for noise reduction.  
    
    \item We evaluate our model's ability to generate programs from partially observable, visually complex demonstrations. We achieve increases of 9\% and 11.75\% for exact and aliased program accuracy relative to previous state-of-the-art. This translates to absolute increases of 5.3\% and 7.7\% respectively and represents the largest single increase in performance on this task to date.

\end{itemize}

%-------------------------------------------------------------------------
\section{Literature Review}
\label{chap5_sec:literature_review}

The task of video understanding \cite{vidU1,vidU2} can be viewed as a subset task of program generation from video demonstrations (PGfVD). As with PGfVD, the understanding of videos requires the ability to extract and understand the correlations between events and features. This is often achieved through models that can perform tasks such as action and perception recognition \cite{kinetics}. However, unlike PGfVD, video understanding aims to describe \textit{what} has been observed in a single demonstration, not \textit{why} something has happened. For the most part, this is a straightforward translation task that can benefit from a large amount of acceptable ambiguity \cite{vid_ambiguity}. This is largely because for the task of video translation, multiple captions may be appropriate and considered semantically correct. However, for the task of PGfVD, the video demonstration may only display a single component of the overall rule set that one is trying to learn. Additionally, the task of PGfVD has much lower levels of ambiguity, as slight changes to a program can result in greatly different outcomes. 

\subsection{Program Induction}
The task of extracting algorithmic representations from observations is known as program induction. Various works have contributed to this field with a diverse set of approaches aimed at solving this problem. \cite{ntm_graves,ntm_gpus,ntm_random} use memory based approaches such as Turing machines while \cite{indcut_dm} adopt end-to-end networks to solve and explain algebraic word problems. However, in contrast to these approaches, we aim to generate a fully defined executable program in a domain specific language. 

\subsection{Program Synthesis}
The aim of program synthesis is to generate a program that captures the underlying logic of given examples. Typically, this task has restricted the programs to simple domain specific languages and has involved producing an abstract syntax tree. Examples of this work include \cite{R3NN}, who proposed using Recursive-Reverse-Recursive neural networks (R3NN) for string transformation. Other work has paired neural models with search algorithms and rule based solvers \cite{deep-coder, PLANS}. Additionally reinforcement learning has also been investigated as a possible way to solve the task of program synthesis \cite{rlnps1, rlnps2}. 

However, most of the work in this field does not consider the task of synthesising programs from visual observations. \cite{demo2program} identified this and proposed the task of generating a program from observing a diverse range of visual demonstrations. To achieve this goal \cite{demo2program} proposed using a sequence-to-sequence LSTM model. Their model consisted of convolutional neural network which fed an LSTM network encoded video frames. \cite{demo2program} introduced a combination of average pooling and a relational network to summarise the encoded demonstrations, which were subsequently passed to a LSTM decoder. \cite{wrc} aimed to improve the computational efficiency of \cite{demo2program} approach by introducing a deviation-pooling summariser to replace the relation network. Additionally, \cite{wrc} proposed using multiple decoding layers to refine the accuracy of the generated program which ultimately resulted in a slight improvement in performance. 

\cite{PLANS} took a different approach to this problem, proposing a hybrid model which combined a neural network to extract specifications and a rule-based solver to generate the program. In particular, \cite{PLANS} proposed using a convolutional neural network to encode the video frames, and two different decoder layers to predict the perceptions and actions observed in the videos. \cite{PLANS} correctly identifies the sensitivity of rule-based solvers to input specification noise \cite{devlin}, and proposed a dynamic filtering method to ignore certain demonstrations based on the confidence level of the neural networks predictions. The inclusion of a dynamic filter allowed \cite{PLANS} to surpass the performance of previous LSTM based program generators. 

%-------------------------------------------------------------------------
\section{Method}
\label{chap5_sec:method}

This section first presents a formal definition for the task of generating programs from video demonstrations, as originally proposed by \cite{demo2program}, before describing the proposed VT4 model, and our contributions in detail. 

\subsection{Program Generation} \label{sec:prog_gen}

The goal of generating a program given $k$ video demonstration can be considered a multi-sequence-to-sequence task. A domain specific language (DSL) is used to define a program which consists of perception primitives, action primitives and control flow statements. Action and perception primitives define the way an agent can interact with and perceive the environment respectively. The control flow statements of the DSL language include while loops, repeat and if/else statements, and simple logic operations. 

\subsubsection{Program}
A program $\pi_\theta(s_t) = a_t$ is defined as a deterministic function, which given an input of state $s \in \mathcal{S}$ at time $t$, returns an action $a \in \mathcal{A}$. For this task we limit the parameters of the program to a vectorised DSL, which we donate as $\theta \in \Theta$. The parameters are what would typically be referred to as the 'code' of the program.

%The code itself is expressed by a tuple of tokens $(w_1, w_2, ..., w_n)$.

\subsubsection{Demonstrations}
A demonstration is defined as a sequence of state-action pairs sampled from $\pi(\theta)$ %consisting of the inputs and outputs (states and actions) of program $\eta$ over time $T$. 

\begin{equation}\label{eq:dem}
    \tau = ((s_1,a_1),(s_2,a_2),...,(s_T,a_T))
\end{equation}

However, there is no guarantee that any particular demonstration generated by an agent following a program $\pi(\theta)$ will provide examples of all control flow statements present in $\pi(\theta)$. Therefore it is necessary to observe a set of demonstrations $\mathcal{D} = (\tau_1, \tau_2, ..., \tau_k)$ where $\mathcal{D} \subset D$ that contains transition examples of all control flow statements in $\pi(\theta)$.

\subsubsection{Actions and Perceptions}
Given the conditional structure of the DSL used to define the parameters $\theta$ for program $\pi$, we employ the use of perception primitives to simplify the high dimensionality of states $s_{1 \rightarrow T}$. Each perception primitive $\mu$ is given as a Boolean value, with $q$ possible perception primitives. This allows for a high level representation of the state to be given by the vector $s=(\mu_1, \mu_2, ..., \mu_q)$.

For an agent interacting in discrete time steps with a deterministic environment of previously defined states $s \in \mathcal{S}$, we can define a finite set of possible actions $\mathcal{A}$. This allows us to model the deterministic transition between states as $\mathcal{P} : \mathcal{S} \times \mathcal{A} \rightarrow \mathcal{S}$. In our case we only have one possible action per transition, and as such we can represent an action primitive as a one hot tensor of length $m$, where $m$ is equal to the total number of possible actions. 

\subsubsection{Visual Language}
Having defined both perception and action primitives, we can use these to define a visual language of semantic tokens $\psi \in \Psi$ with a deterministic function $F(x,y)$ such that,

\begin{equation}
    \psi_t = F(s_t, a_t) 
\end{equation}

This approach of tokenizing high level semantic information for ease of use with transformer models has been shown to be very effective by \cite{vidbert}. With this we can also substitute our transition tokens into equation \ref{eq:dem} giving,

\begin{equation}
    \tau = (\psi_1, \psi_2, ... \psi_T)
\end{equation}

%\subsubsection{Supervision Signals}
%We abide by the constraints originally proposed by \cite{demo2program} and assume that action and perception labels are absent during testing. However, we form the same conclusion as \cite{demo2program, wrc, PLANS} that during training the action and perception signals are required for the model to learn how to generate  the I/O specifications of the program. 

\subsection{VT4 Model}
Our VT4 model can be separated into two main sections; i) the semantic encoder and ii) the program generator network. We approach this problem of generating an executable program from video demonstrations as a combined translation and summarisation task. In contrast to previous approaches which simplify the problem to a straight sequence-to-sequence task, (by creating a summarised expression of the multiple demonstrations) our approach frames the problem as the multi-sequence-to-sequence task that it is. This eliminates the inherent probability of information loss associated with summarising multiple diverse demonstrations.

\subsubsection{Semantic Encoder}\label{semantic encoder}
 The Semantic Encoder itself can be separated into two parts: a neural module and a tokenizer. The neural module is a multi-layer convolutional neural network (CNN) which learns to detect the perception primitives present in each frame, and the actions taken between frames. The action prediction network consists of a five-layer convolutional neural network with two fully connected layers, and is trained from scratch. The perception prediction network utilises a pre-trained Efficientnet model \cite{effnet}. 
 
 Given a set of video demonstrations $\mathcal{V} = \{v_i\}^k_{i=1}$, we desire the corresponding perceptions $p$ and actions $a$ sequences for each demonstration. This is a straight forward problem which is modelled as;
 \begin{equation}
 p_{i,j} = MLP(CNN(v_{i,j})) 
 \end{equation}
 \begin{equation}
  a_{i,j} = MLP(CNN(v_{i,j}), CNN(v_{i,j+1}))
\label{eq:act}
\end{equation}
 
Where $i$ and $j$ refer to the $i^{th}$ demonstration and $j^{th}$ frame. While it is possible to predict all the observable perceptions from a single frame $\boldsymbol{v}_{i,j}$ with purely spatial information, this is not the case for predicting actions. To predict the action taken at any time $t$ temporal information in the form of a minimum of two frames is required. Thus, we concatenated sequential frames together as shown in equation \ref{eq:act} for action prediction. 
 
Having now obtained the predicted actions $a_{i,j}$ and perceptions $\boldsymbol{p}_{i,j}$ for each frame of every video, we are able to use these to create semantic tokens which completely encapsulates all the required information from each frame. We achieve this by passing our predictions through the second part of the semantic encoder which we refer to as a tokenizer. The tokenizer itself is a deterministic function that takes as input the predicted action and perceptions of each frame individually.  

\begin{equation}
\psi_{i,j} = F(p_{i,j}, a_{i,j}) 
\label{eq:token2}
\end{equation}

As described in section \ref{sec:prog_gen}, our perception prediction is a multiclass prediction problem given as $p_{i,j} = (\mu_1, \mu_2, ..., \mu_q)$, while our action prediction returns a single class prediction. Our tokenizer first concatenates our predictions into a single tensor $pa$, before calculating the finite sum of a basic power series.  

\begin{equation}
\psi_{i,j} = \sum_{n=0}^n pa_n \times 2^n
\label{eq:tok_func}
\end{equation}Our demonstrations are now encoded into a sequence of semantic tokens which encapsulates all the semantic information derived from the original video frames. 

We pass every tokenised demonstrations into our program generator network as a set of disjoint sequences. This approach of creating 'visual words'\cite{vidbert} allows us to take inspiration from the field of natural language processing and concatenate our demonstrations together separating the different demonstrations with special tokens. $V = (<start>, \psi_{1,1}, \psi_{1,2} ..., \psi_{1,l} <sep>, ..., \psi_{k,l},  <end>)$

\begin{figure*}
    \includegraphics[width=\textwidth,]{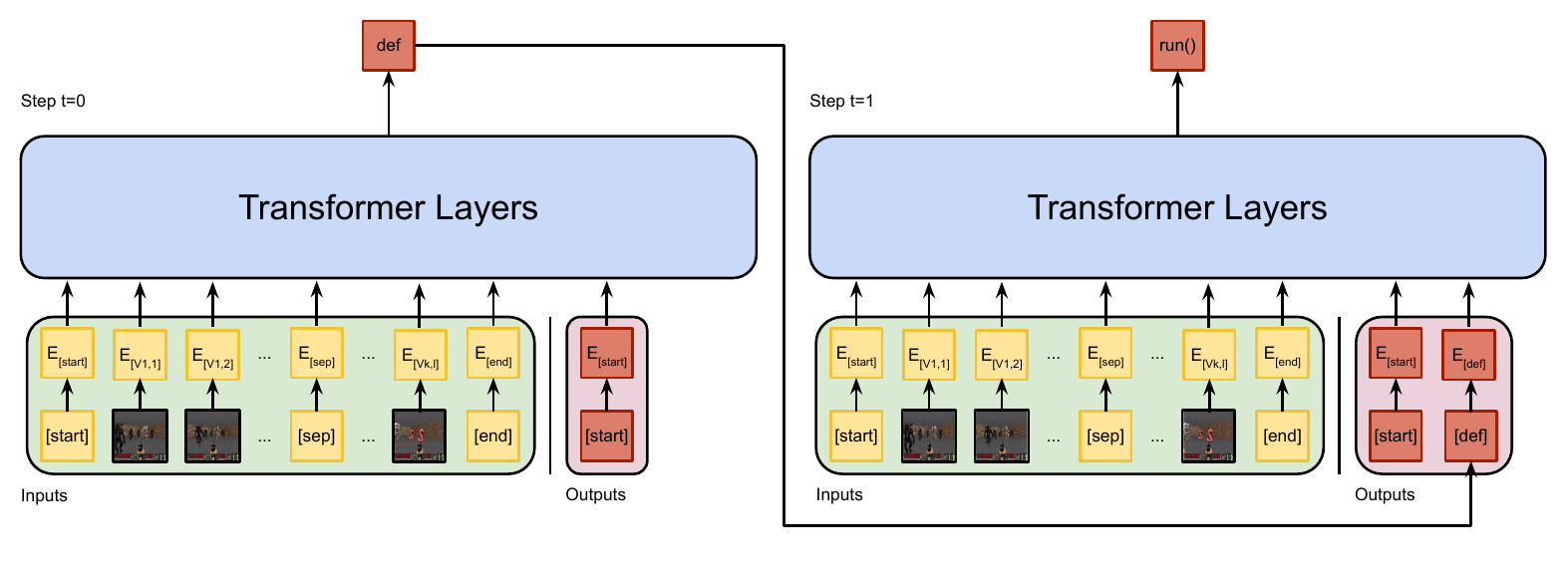}
    \caption[A diagram of the Program Generator Module]{The Program Generator Module. The program generator consists of an encoder-decoder transformer network which receives as input the visual language tokens from the semantic encoder. The network iteratively decodes the program in a sequential fashion.}
    \label{fig:T5}
\end{figure*} 

\subsubsection{Program Generator Network}

Due to the wide success of Transformer based architectures \cite{transformer} in various settings such as machine translation \cite{MTT1,MTT2}, natural language processing \cite{T5,bert}, and even more recently image classification \cite{img_trans}, we hypothesis that they should also be highly effective for the task of program generation. As such our program generator is an encoder-decoder transformer network. By utilising the insights of \cite{vidbert} and converting our inputs into a visual language, we can leverage pre-trained language models. We choose to leverage a pre-trained 'Text-to-Text Transfer Transformer' (T5) network as proposed by \cite{T5}.

The T5 model is particularly well suited to our task as it specifically casts all tasks as a text-to-text task. Due to this we can leverage pre-trained features from upstream natural language tasks such as summarisation and translation. The T5 implementation closely follows the originally proposed architecture of \cite{transformer}. The encoder consists of a stack of 'blocks', wherein each block is comprised of a self-attention, layer normalisation, and a feed-forward network. The decoder has a similar structure, except that it includes an additional standard attention mechanism. This standard attention mechanism is applied after the self-attention layer and attends to the output from the encoder. Figure \ref{fig:T5} gives an overview of generative sequence of the transformer network, while for full details we refer the reader to \cite{T5} and \cite{transformer} respectively.

%-------------------------------------------------------------------------
\section{Experiments}
\label{chap5_sec:experiments}

In this section we present the experimental evaluation of our model. We include an overview of the dataset and metrics used for this evaluation. We then discuss the results of our experiments along with providing a noise-ablation study to evaluate our model's ability to handle noisy inputs.  

\subsection{Dataset and Metrics}
\subsubsection{Dataset}
We evaluate our model with the Vizdoom Program Dataset\cite{demo2program} which has the following structure. For every program label there exists multiple video demonstrations of an agent following said program in the deterministic virtual environment known as Vizdoom\cite{vizdoom}. For every demonstration (of length $T$), there exists action and perception labels of lengths $T-1$ and $T$ respectively. We conform with the previously established convention set by \cite{demo2program} and utilise a total of twenty-five demonstrations per program during testing and training. We also adhere to the assumption that action and perception labels are only available during training, and that at test time we only have access to the video demonstrations of the agent. The dataset contains 80,000 programs for training, and 8,000 programs for testing. 

%\subsubsection{Metrics}
%As the problem of verifying that two programs are in fact equal is an intractable problem, we evaluate the accuracy of our model by comparing the synthesised parameters $\hat{\theta}$ with the instantiated parameters of the ground truth program $\theta^*$. In particular, we evaluate the \emph{exact accuracy} and the \emph{aliased accuracy}. 

\subsubsection{Aliased and Exact Accuracy}
As the problem of verifying that two programs are in fact equal is an intractable problem, we evaluate the accuracy of our model by comparing the synthesised parameters $\hat{\theta}$ with the instantiated parameters of the ground truth program $\theta^*$. We consider a program to be an exact match if, and only if, the synthesised parameters $\hat{\theta}$ is an exact match to that of the instantiated ground truth parameters $\theta^*$. This is formally expressed as,

\begin{equation}
    Acc_{exact} = \frac{1}{N}\sum^N_{n=1}1_{exact}(\hat{\theta},\theta^*)
\end{equation}

%\subsubsection{Aliased Accuracy}
While the exact accuracy is a simplistic measure of the performance of our model, it does not account for the ambiguity of the program space. As identified by \cite{demo2program}, it is possible to exploit the simplistic syntax of our DSL and enumerate multiple variations of the code following a set of defined rules. Examples of this include decomposing control flow statements such as if/else statements, and unfolding repeat statements. With this we can formally express the aliased accuracy as

\begin{equation}
    Acc_{alias} = \frac{1}{N}\sum^N_{n=1}1_{alias}(\hat{\theta},\theta^*)
\end{equation}

\subsection{Overall Performance}

We report the results of our evaluation on the Vizdoom benchmark in table \ref{tab:main_acc}. We strictly adhere to the experimental settings originally proposed by \cite{demo2program} to provide a fair comparison. Our VT4 model significantly improves on the exact and aliased program accuracy of the prior state-of-the-art with relative 9\% and 11.75\% increases and 5.3\% and 7.7\% absolute increases respectively. These results provide clear evidence of the capabilities of our model to simultaneously perform summarisation and translation of disjointed sequences. These results also provide empirical evidence to support the use of visual languages to describe the semantics of complex scenes for tasks requiring translation. Additionally, we observe the expected result of higher accuracy for aliased programs compared to exact programs as consistently observed across all prior works. 

%We also investigate claims by \cite{wrc} and \cite{PLANS} that the original algorithms used by \cite{demo2program} fails to recognise all semantically identical programs. In particular are cases such as the example shown in figure \ref{fig:gt_vs_syn}. While we find these claims to be accurate, we report our results using the same metrics to keep our comparison fair. However, we strongly agree with \cite{PLANS} on the need to improve this metric in future work. 

%\begin{figure}
%    \begin{center}
%    \includegraphics[height=0.5\linewidth]{images/program_examples.pdf}
%    \caption[A comparison of the ground truth program and a synthesised program
%from our VT4 model.]{A comparison of the ground truth program and a synthesised program from our VT4 model. While these two programs are semantically equal, the algorithm used by \cite{demo2program} considers these as different. To keep evaluation fair we report results using the same base code wherein this situation is considered a 'failure' case.}
%    \Description{DSL}
%    \label{fig:gt_vs_syn}
%    \end{center}
%\end{figure}

\begin{table}[]
    \centering
    \begin{tabular}{lll}
    Model             & Exact & Aliased \\ \hline\hline
    demo2program\cite{demo2program}      & 53.2     & 62.5       \\
    watch-reason-code\cite{wrc} & 55.8    & 63.4      \\
    PLANS (dynamic)\cite{PLANS}  & 58.8 $\pm{0.6}$    & 65.5 $\pm{0.6}$      \\
    VT4 (ours)        & \textbf{64.1}    & \textbf{73.2} \\ \hline
    \end{tabular}
    \caption{An exact comparison of our results compared to the results of previously published works. }
    \label{tab:main_acc}
\end{table}
\subsection{Noise Ablation Study}
While our model is clearly capable of inductive reasoning over multiple demonstrations, we consider the implication of noise with respect to its ability to accurately generate programs. Previous approaches that utilise rule-based solvers \cite{PLANS} have been highly sensitive to noise. In-fact, even with high levels of perception accuracy the PLANS model required dynamic filtering. This indicates that even the slightest amount noise in the input causes the PLANS model to underperform. 

To test our model's ability to deal with noise in its input we devise a noise ablation study. By separately predicting the actions and perceptions with two distinct networks we can vary the accuracy of the perception predictions independently of the actions. Our semantic encoder in this setup has independent action and perception encoders. Our action encoder easily obtains an accuracy of 98\% with a simple five layer convolutional neural network (as described in section \ref{semantic encoder}). We train a perception encoder with the same architecture which achieves an accuracy of 79.9\%. We then utilise a pre-trained object detection network (efficientnet \cite{effnet}) to improve this result. This time our prediction encoder achieves an accuracy of 90.2\%. Our approach to utilise predicted perception primitives and actions to generate a visual language allows us to evaluate a 'perfect case' scenario. As we have access to ground-truth labels for these perception primitives and actions, we can use these directly to generate our visual language tokens. Doing so artificially creates a scenario in which our model has effectively no input noise. 

Our results from this study are presented in table \ref{tab:per_acc}. These results clearly show our models capacity to not only reason over multiple demonstrations, but to handle contradictory or noisy signals. In contrast to rule-based solvers, which as shown by \cite{PLANS} are strongly reliant on pre-defined filtering heuristics, our VT4 model is able to learn its own heuristics. These results also show that our model was able to exceed the previous state-of-the-art while enduring a 20\% error rate in perception predictions.

\begin{table}
	\centering
	\begin{tabular}{lll}
	\hline
	Perception Noise & Exact & Alias \\ \hline\hline
  0\% &	66.0 & 75.1 \\
  10\% &	64.1 & 73.2 \\
  20\% &	62.7 & 71.6 \\ \hline
  
	\end{tabular}
    \caption{Exact and Alias program accuracy's for varying levels of perception noise. }
    \label{tab:per_acc}
\end{table}

%-------------------------------------------------------------------------
\section{Conclusions}
The task of synthesising a program from multiple visual demonstrations involves solving many different tasks. These include learning spatial-temporal relationship from visually complex inputs and successfully translating these into a logical sequence in a different domain. Previous attempts at solving this problem have relied upon summarised representations of the spatial-temporal relationships and have been highly sensitive to input noise. In this paper we propose a video-to-text transfer transformer network that is able to perform multi-sequence-to-sequence translation without requiring summarised spatial-temporal embeddings. Our method is also highly robust to input noise, which is a problem that caused significant challenges for previous methods. On top of this, we achieve the largest increase in performance to date on this task.

%-------------------------------------------------------------------------

\bibliography{egbib}

\begin{thebibliography}{29}
\providecommand{\natexlab}[1]{#1}
\providecommand{\url}[1]{\texttt{#1}}
\expandafter\ifx\csname urlstyle\endcsname\relax
  \providecommand{\doi}[1]{doi: #1}\else
  \providecommand{\doi}{doi: \begingroup \urlstyle{rm}\Url}\fi

\bibitem[Ahmed et~al.(2017)Ahmed, Keskar, and Socher]{MTT2}
Karim Ahmed, Nitish~Shirish Keskar, and Richard Socher.
\newblock Weighted transformer network for machine translation.
\newblock \emph{CoRR}, abs/1711.02132, 2017.
\newblock URL \url{http://arxiv.org/abs/1711.02132}.

\bibitem[{Alur} et~al.(2013){Alur}, {Bodik}, {Juniwal}, {Martin},
  {Raghothaman}, {Seshia}, {Singh}, {Solar-Lezama}, {Torlak}, and
  {Udupa}]{syntax_guided}
R.~{Alur}, R.~{Bodik}, G.~{Juniwal}, M.~M.~K. {Martin}, M.~{Raghothaman}, S.~A.
  {Seshia}, R.~{Singh}, A.~{Solar-Lezama}, E.~{Torlak}, and A.~{Udupa}.
\newblock Syntax-guided synthesis.
\newblock In \emph{2013 Formal Methods in Computer-Aided Design}, pages 1--8,
  2013.
\newblock \doi{10.1109/FMCAD.2013.6679385}.

\bibitem[Balog et~al.(2016)Balog, Gaunt, Brockschmidt, Nowozin, and
  Tarlow]{deep-coder}
Matej Balog, Alexander Gaunt, Marc Brockschmidt, Sebastian Nowozin, and Daniel
  Tarlow.
\newblock Deepcoder: Learning to write programs.
\newblock 11 2016.

\bibitem[Bunel et~al.(2018)Bunel, Hausknecht, Devlin, Singh, and Kohli]{rlnps1}
Rudy Bunel, Matthew~J. Hausknecht, Jacob Devlin, Rishabh Singh, and Pushmeet
  Kohli.
\newblock Leveraging grammar and reinforcement learning for neural program
  synthesis.
\newblock \emph{CoRR}, abs/1805.04276, 2018.
\newblock URL \url{http://arxiv.org/abs/1805.04276}.

\bibitem[Dang-Nhu(2020)]{PLANS}
Rapha{\"e}l Dang-Nhu.
\newblock Plans: Robust program learning from neurally inferred specifications.
\newblock \emph{ArXiv}, abs/2006.03312, 2020.

\bibitem[Devlin et~al.(2017)Devlin, Uesato, Bhupatiraju, Singh, Mohamed, and
  Kohli]{devlin}
Jacob Devlin, Jonathan Uesato, Surya Bhupatiraju, Rishabh Singh, Abdel-rahman
  Mohamed, and Pushmeet Kohli.
\newblock Robustfill: Neural program learning under noisy i/o.
\newblock In \emph{Proceedings of the 34th International Conference on Machine
  Learning - Volume 70}, ICML'17, page 990–998. JMLR.org, 2017.

\bibitem[Devlin et~al.(2018)Devlin, Chang, Lee, and Toutanova]{bert}
Jacob Devlin, Ming{-}Wei Chang, Kenton Lee, and Kristina Toutanova.
\newblock {BERT:} pre-training of deep bidirectional transformers for language
  understanding.
\newblock \emph{CoRR}, abs/1810.04805, 2018.
\newblock URL \url{http://arxiv.org/abs/1810.04805}.

\bibitem[Dong et~al.(2019)Dong, Gao, Chen, Guo, Cao, and Zhang]{vid_ambiguity}
Jiarong Dong, Ke~Gao, Xiaokai Chen, Junbo Guo, Juan Cao, and Yongdong Zhang.
\newblock Not all words are equal: Video-specific information loss for video
  captioning.
\newblock \emph{CoRR}, abs/1901.00097, 2019.
\newblock URL \url{http://arxiv.org/abs/1901.00097}.

\bibitem[Dosovitskiy et~al.(2020)Dosovitskiy, Beyer, Kolesnikov, Weissenborn,
  Zhai, Unterthiner, Dehghani, Minderer, Heigold, Gelly, Uszkoreit, and
  Houlsby]{img_trans}
Alexey Dosovitskiy, Lucas Beyer, Alexander Kolesnikov, Dirk Weissenborn,
  Xiaohua Zhai, Thomas Unterthiner, Mostafa Dehghani, Matthias Minderer, Georg
  Heigold, Sylvain Gelly, Jakob Uszkoreit, and Neil Houlsby.
\newblock An image is worth 16x16 words: Transformers for image recognition at
  scale, 2020.

\bibitem[Duan et~al.(2019)Duan, Wu, Gan, Zhang, Huang, van~den Hengel, and
  Zhu]{wrc}
Xuguang Duan, Qi~Wu, Chuang Gan, Yiwei Zhang, Wenbing Huang, Anton van~den
  Hengel, and Wenwu Zhu.
\newblock Watch, reason and code: Learning to represent videos using program.
\newblock In \emph{Proceedings of the 27th ACM International Conference on
  Multimedia}, MM '19, page 1543–1551, New York, NY, USA, 2019. Association
  for Computing Machinery.
\newblock ISBN 9781450368896.
\newblock \doi{10.1145/3343031.3351094}.
\newblock URL \url{https://doi.org/10.1145/3343031.3351094}.

\bibitem[Graves et~al.(2014)Graves, Wayne, and Danihelka]{ntm_graves}
Alex Graves, Greg Wayne, and Ivo Danihelka.
\newblock Neural turing machines.
\newblock \emph{CoRR}, abs/1410.5401, 2014.
\newblock URL \url{http://arxiv.org/abs/1410.5401}.

\bibitem[Gulwani(2011)]{ai_dream2}
Sumit Gulwani.
\newblock Automating string processing in spreadsheets using input-output
  examples.
\newblock \emph{SIGPLAN Not.}, 46\penalty0 (1):\penalty0 317–330, January
  2011.
\newblock ISSN 0362-1340.
\newblock \doi{10.1145/1925844.1926423}.
\newblock URL \url{https://doi.org/10.1145/1925844.1926423}.

\bibitem[Jha et~al.(2010)Jha, Gulwani, Seshia, and Tiwari]{oracle_specs}
Susmit Jha, Sumit Gulwani, Sanjit~A. Seshia, and Ashish Tiwari.
\newblock Oracle-guided component-based program synthesis.
\newblock ICSE '10, page 215–224, New York, NY, USA, 2010. Association for
  Computing Machinery.
\newblock ISBN 9781605587196.
\newblock \doi{10.1145/1806799.1806833}.
\newblock URL \url{https://doi.org/10.1145/1806799.1806833}.

\bibitem[Kaiser and Sutskever(2016)]{ntm_gpus}
Lukasz Kaiser and Ilya Sutskever.
\newblock Neural gpus learn algorithms.
\newblock In Yoshua Bengio and Yann LeCun, editors, \emph{4th International
  Conference on Learning Representations, {ICLR} 2016, San Juan, Puerto Rico,
  May 2-4, 2016, Conference Track Proceedings}, 2016.
\newblock URL \url{http://arxiv.org/abs/1511.08228}.

\bibitem[Kay et~al.(2017)Kay, Carreira, Simonyan, Zhang, Hillier,
  Vijayanarasimhan, Viola, Green, Back, Natsev, Suleyman, and
  Zisserman]{kinetics}
Will Kay, Jo{\~{a}}o Carreira, Karen Simonyan, Brian Zhang, Chloe Hillier,
  Sudheendra Vijayanarasimhan, Fabio Viola, Tim Green, Trevor Back, Paul
  Natsev, Mustafa Suleyman, and Andrew Zisserman.
\newblock The kinetics human action video dataset.
\newblock \emph{CoRR}, abs/1705.06950, 2017.
\newblock URL \url{http://arxiv.org/abs/1705.06950}.

\bibitem[Kempka et~al.(2016)Kempka, Wydmuch, Runc, Toczek, and
  Jaskowski]{vizdoom}
Michal Kempka, Marek Wydmuch, Grzegorz Runc, Jakub Toczek, and Wojciech
  Jaskowski.
\newblock Vizdoom: {A} doom-based {AI} research platform for visual
  reinforcement learning.
\newblock \emph{CoRR}, abs/1605.02097, 2016.
\newblock URL \url{http://arxiv.org/abs/1605.02097}.

\bibitem[Kurach et~al.(2016)Kurach, Andrychowicz, and Sutskever]{ntm_random}
Karol Kurach, Marcin Andrychowicz, and Ilya Sutskever.
\newblock Neural random-access machines.
\newblock In Yoshua Bengio and Yann LeCun, editors, \emph{4th International
  Conference on Learning Representations, {ICLR} 2016, San Juan, Puerto Rico,
  May 2-4, 2016, Conference Track Proceedings}, 2016.
\newblock URL \url{http://arxiv.org/abs/1511.06392}.

\bibitem[Lin et~al.(2019)Lin, Gan, and Han]{vidU1}
Ji~Lin, Chuang Gan, and Song Han.
\newblock Tsm: Temporal shift module for efficient video understanding.
\newblock In \emph{Proceedings of the IEEE/CVF International Conference on
  Computer Vision (ICCV)}, October 2019.

\bibitem[Ling et~al.(2017)Ling, Yogatama, Dyer, and Blunsom]{indcut_dm}
Wang Ling, Dani Yogatama, Chris Dyer, and Phil Blunsom.
\newblock Program induction by rationale generation: Learning to solve and
  explain algebraic word problems.
\newblock \emph{CoRR}, abs/1705.04146, 2017.
\newblock URL \url{http://arxiv.org/abs/1705.04146}.

\bibitem[Parisotto et~al.(2016)Parisotto, Mohamed, Singh, Li, Zhou, and
  Kohli]{R3NN}
Emilio Parisotto, Abdel{-}rahman Mohamed, Rishabh Singh, Lihong Li, Dengyong
  Zhou, and Pushmeet Kohli.
\newblock Neuro-symbolic program synthesis.
\newblock \emph{CoRR}, abs/1611.01855, 2016.
\newblock URL \url{http://arxiv.org/abs/1611.01855}.

\bibitem[Raffel et~al.(2019)Raffel, Shazeer, Roberts, Lee, Narang, Matena,
  Zhou, Li, and Liu]{T5}
Colin Raffel, Noam Shazeer, Adam Roberts, Katherine Lee, Sharan Narang, Michael
  Matena, Yanqi Zhou, Wei Li, and Peter~J. Liu.
\newblock Exploring the limits of transfer learning with a unified text-to-text
  transformer.
\newblock \emph{CoRR}, abs/1910.10683, 2019.
\newblock URL \url{http://arxiv.org/abs/1910.10683}.

\bibitem[Simmons{-}Edler et~al.(2018)Simmons{-}Edler, Miltner, and
  Seung]{rlnps2}
Riley Simmons{-}Edler, Anders Miltner, and H.~Sebastian Seung.
\newblock Program synthesis through reinforcement learning guided tree search.
\newblock \emph{CoRR}, abs/1806.02932, 2018.
\newblock URL \url{http://arxiv.org/abs/1806.02932}.

\bibitem[{Sun} et~al.(2019){Sun}, {Myers}, {Vondrick}, {Murphy}, and
  {Schmid}]{vidbert}
C.~{Sun}, A.~{Myers}, C.~{Vondrick}, K.~{Murphy}, and C.~{Schmid}.
\newblock Videobert: A joint model for video and language representation
  learning.
\newblock In \emph{2019 IEEE/CVF International Conference on Computer Vision
  (ICCV)}, pages 7463--7472, 2019.
\newblock \doi{10.1109/ICCV.2019.00756}.

\bibitem[Sun et~al.(2018)Sun, Noh, Somasundaram, and Lim]{demo2program}
Shao-Hua Sun, Hyeonwoo Noh, Sriram Somasundaram, and Joseph Lim.
\newblock Neural program synthesis from diverse demonstration videos.
\newblock In Jennifer Dy and Andreas Krause, editors, \emph{Proceedings of the
  35th International Conference on Machine Learning}, volume~80 of
  \emph{Proceedings of Machine Learning Research}, pages 4790--4799. PMLR,
  10--15 Jul 2018.
\newblock URL \url{http://proceedings.mlr.press/v80/sun18a.html}.

\bibitem[Tan and Le(2019)]{effnet}
Mingxing Tan and Quoc Le.
\newblock {E}fficient{N}et: Rethinking model scaling for convolutional neural
  networks.
\newblock In Kamalika Chaudhuri and Ruslan Salakhutdinov, editors,
  \emph{Proceedings of the 36th International Conference on Machine Learning},
  volume~97 of \emph{Proceedings of Machine Learning Research}, pages
  6105--6114. PMLR, 09--15 Jun 2019.
\newblock URL \url{http://proceedings.mlr.press/v97/tan19a.html}.

\bibitem[Vaswani et~al.(2017)Vaswani, Shazeer, Parmar, Uszkoreit, Jones, Gomez,
  Kaiser, and Polosukhin]{transformer}
Ashish Vaswani, Noam Shazeer, Niki Parmar, Jakob Uszkoreit, Llion Jones,
  Aidan~N. Gomez, undefinedukasz Kaiser, and Illia Polosukhin.
\newblock Attention is all you need.
\newblock In \emph{Proceedings of the 31st International Conference on Neural
  Information Processing Systems}, NIPS'17, page 6000–6010, Red Hook, NY,
  USA, 2017. Curran Associates Inc.
\newblock ISBN 9781510860964.

\bibitem[Waldinger and Lee(1969)]{ai_dream1}
Richard~J. Waldinger and Richard C.~T. Lee.
\newblock Prow: A step toward automatic program writing.
\newblock In \emph{Proceedings of the 1st International Joint Conference on
  Artificial Intelligence}, IJCAI'69, page 241–252, San Francisco, CA, USA,
  1969. Morgan Kaufmann Publishers Inc.

\bibitem[Wang et~al.(2019)Wang, Li, Xiao, Zhu, Li, Wong, and Chao]{MTT1}
Qiang Wang, Bei Li, Tong Xiao, Jingbo Zhu, Changliang Li, Derek~F. Wong, and
  Lidia~S. Chao.
\newblock Learning deep transformer models for machine translation.
\newblock \emph{CoRR}, abs/1906.01787, 2019.
\newblock URL \url{http://arxiv.org/abs/1906.01787}.

\bibitem[Wu et~al.(2019)Wu, Feichtenhofer, Fan, He, Krahenbuhl, and
  Girshick]{vidU2}
Chao-Yuan Wu, Christoph Feichtenhofer, Haoqi Fan, Kaiming He, Philipp
  Krahenbuhl, and Ross Girshick.
\newblock Long-term feature banks for detailed video understanding.
\newblock In \emph{Proceedings of the IEEE/CVF Conference on Computer Vision
  and Pattern Recognition (CVPR)}, June 2019.

\end{thebibliography}
\end{document}